\documentclass[11pt,a4paper]{article}
\usepackage[hyperref]{acl2020}
\usepackage{times}
\usepackage{latexsym}

\usepackage{graphicx}

\usepackage{amsmath, amssymb}
\usepackage{color}
\usepackage{xcolor}
\usepackage{booktabs}
\usepackage{algorithm}
\usepackage[noend]{algpseudocode}
\usepackage{amssymb}
\usepackage{pifont}
\usepackage{lipsum}
\usepackage{capt-of}
\usepackage{cancel}
\usepackage{microtype}

\aclfinalcopy 

\title{Global Locality in Biomedical Relation and Event Extraction}

\author{Elaheh ShafieiBavani, Antonio Jimeno Yepes, Xu Zhong, David Martinez Iraola \\
	IBM Research Australia\\
	\texttt{\{elaheh.shafieibavani, david.martinez.iraola1\}@ibm.com}\\ \texttt{\{antonio.jimeno, peter.zhong\}@au1.ibm.com}\\}

\date{}

\begin{document}
\maketitle
\begin{abstract}
	Due to the exponential growth of biomedical literature, event and relation extraction are important tasks in biomedical text mining. Most work only focus on relation extraction, and detect a single entity pair mention on a short span of text, which is not ideal due to long sentences that appear in biomedical contexts. We propose an approach to both relation and event extraction, for simultaneously predicting relationships between all mention pairs in a text. We also perform an empirical study to discuss different network setups for this purpose. The best performing model includes a set of multi-head attentions and convolutions, an adaptation of the transformer architecture, which offers self-attention the ability to strengthen dependencies among related elements, and models the interaction between features extracted by multiple attention heads. Experiment results demonstrate that our approach outperforms the state of the art on a set of benchmark biomedical corpora including BioNLP 2009, 2011, 2013 and BioCreative 2017 shared tasks. 
\end{abstract}

\section{Introduction}\label{sec:intro}

Event and relation extraction has become a key research topic in natural language processing with a variety of practical applications especially in the biomedical domain, where these methods are widely used to extract information from massive document sets, such as scientific literature and patient records. This information contains the interactions between named entities such as protein-protein, drug-drug, chemical-disease, and more complex events. 

\begin{figure}[h!]
	\centering
	\includegraphics[scale=0.26]{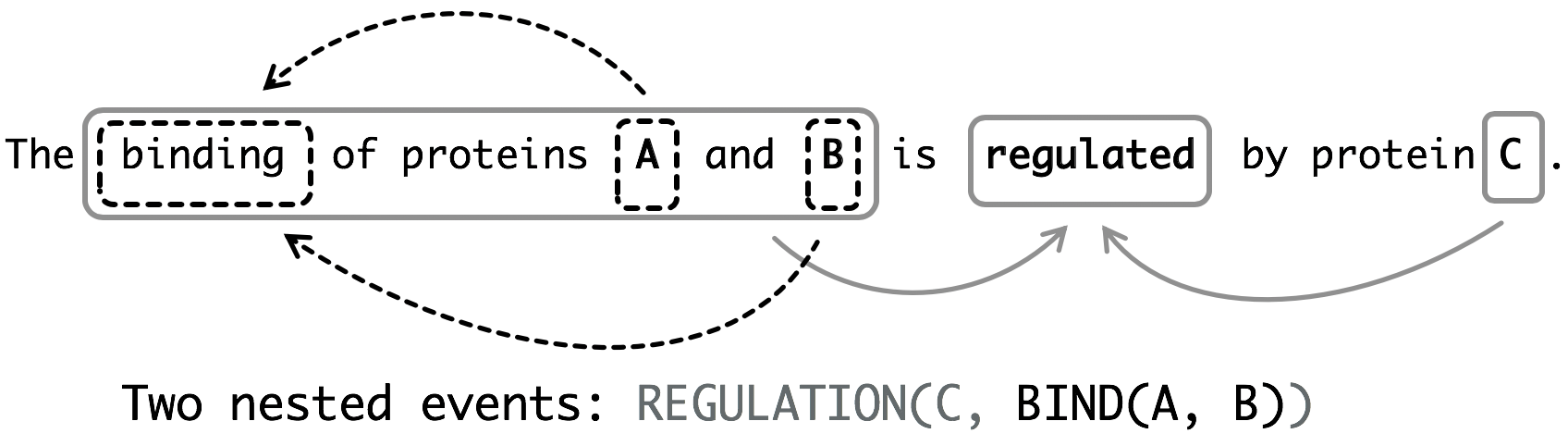}
	\caption{Example of nested events from GE11 shared task}
	\label{fig:example}
\end{figure}

Relations are usually described as typed, sometimes directed, pairwise links between defined
named entities \cite{bjorne2009extracting}. Event extraction differs from relation extraction in the sense that an event has an annotated trigger word (e.g., a verb), and could be an argument of other events to connect more than two entities. Event extraction is a more complicated task compared to relation extraction due to the tendency of events to capture the semantics of texts. For clarity, Figure \ref{fig:example} shows an example from the GE11 shared task corpus that includes two nested events.

Recently, deep neural network models obtain state-of-the-art performance for event and relation extraction. Two major neural network architectures for this purpose include Convolutional Neural Networks (CNNs) \cite{santos2015classifying,zeng2015distant} and Recurrent Neural Networks (RNNs) \cite{mallory2015large,verga2015multilingual,zhou2016attention}. While CNNs can capture the local features based on the convolution operations and are more suitable for addressing short sentence sequences, RNNs are good at learning long-term dependency features, which are considered more suitable for dealing with long sentences. Therefore, combining the advantages of both models is the key point for improving biomedical event and relation extraction performance \cite{zhang2018hybrid}.

However, encoding long sequences to incorporate long-distance context is very expensive in RNNs \cite{verga2018simultaneously} due to their computational dependence on the length of the sequence. In addition, computations could not be parallelized since each token’s representation requires as input the representation of its previous token. In contrast, CNNs can be executed entirely in parallel across the sequence, and have shown good performance in event and relation extraction \cite{bjorne2018biomedical}. However, the amount of context incorporated into a single token’s representation is limited by the depth of the network, and very deep networks can be difficult to learn \cite{hochreiter1998vanishing}. 

To address these problems, self-attention networks \cite{parikh2016decomposable,lin2017structured} come into play. They have shown promising empirical results in various natural language processing tasks, such as information extraction \cite{verga2018simultaneously}, machine translation \cite{vaswani2017attention} and natural language inference \cite{shen2018disan}. One of their strengths lies in their high parallelization in computation and flexibility in modeling dependencies regardless of distance by explicitly attending to all the elements. In addition, their performance can be improved by multi-head attention \cite{vaswani2017attention}, which projects the input sequence into multiple subspaces and applies attention to the representation in each subspace.

\begin{figure*}[h!]
	\centering
	\includegraphics[scale=0.48]{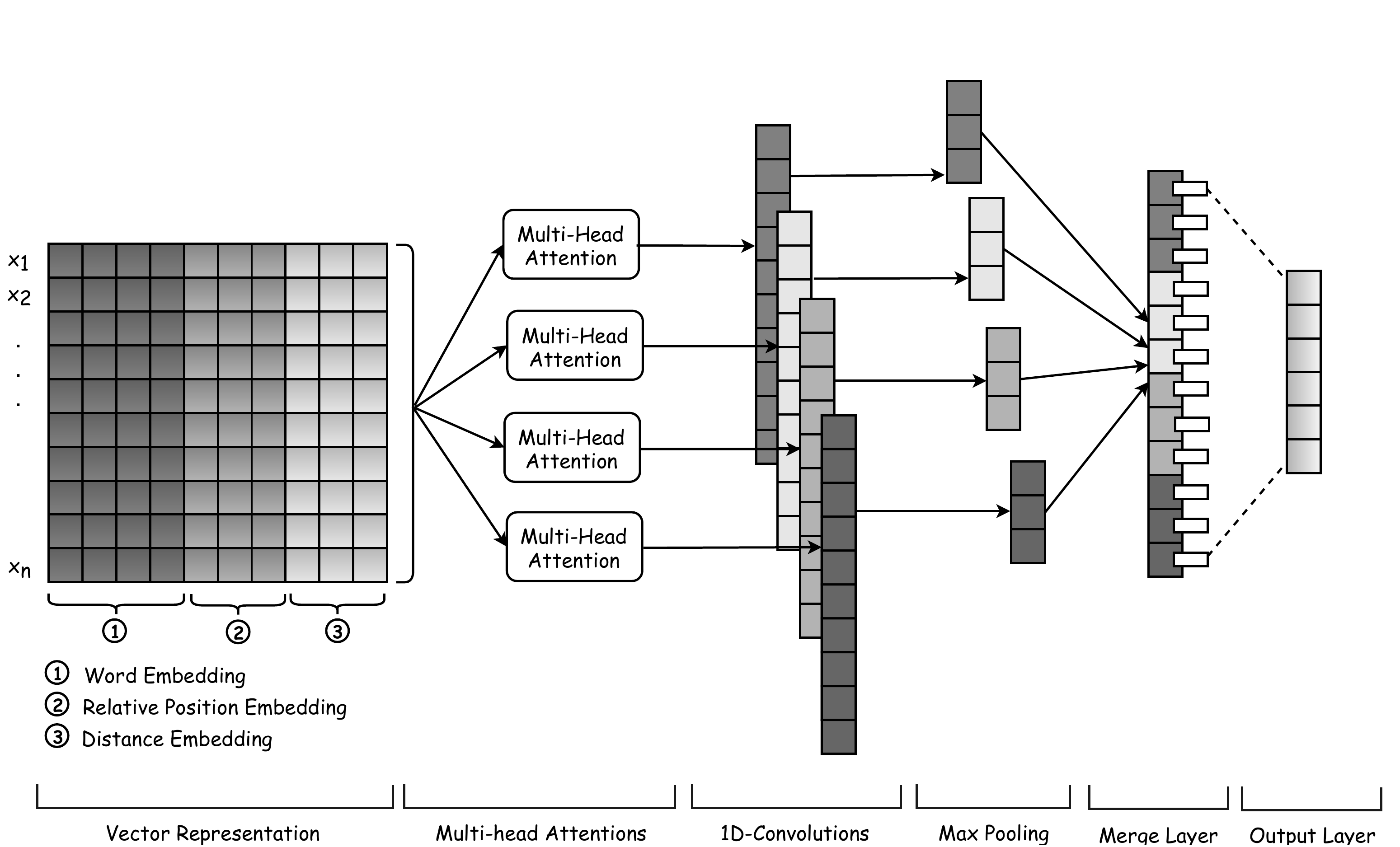}
	\caption{Our model architecture for biomedical event and relation extraction: The embedding vectors
		are merged together before the multi-head attention and convolution layers. The global max pooling is then applied to the results of these operations. Finally, the output layer shows the predicted labels.}
	\label{fig:model}
\end{figure*}

In this paper, we propose a new neural network model 
that combines multi-head attention mechanisms with a set of convolutions to provide global locality in biomedical event and relation extraction. Convolutions capture the local structure of text, while self-attention learns the global interaction between each pair of words. Hence, our approach models locality for self-attention while the interactions between features are learned by multi-head attentions. The experiment results over the biomedical benchmark corpora show that providing global locality outperforms the existing state of the art for biomedical event and relation extraction. The proposed architecture is shown in Figure~\ref{fig:model}.

Conducting a set of experiments over the corpora of the shared tasks for BioNLP 2009, 2011 and 2013, and BioCreative 2017, we compare the performance of our model with the best-performing system (TEES) \cite{bjorne2018biomedical} in the shared tasks. The results we achieve via precision, recall, and F-score demonstrate that our model obtains state-of-the-art performance. We also empirically assess three variants of our model and elaborate on the results further in the experiments.

The rest of the paper is organized as follows. Section \ref{sec:bkg} summarizes the background. The data, and the proposed approach are explained in Sections \ref{sec:data} and \ref{sec:model} respectively. Section \ref{sec:expr} explains the experiments and discusses the achieved results. Finally, Section \ref{sec:concl} summarizes the findings of the paper and presents future work.

\section{Background}\label{sec:bkg}

Biomedical event and relation extraction have been developed thanks to the contribution of corpora generated for community shared tasks \cite{kim2009overview,kim2011overview,nedellec2013overview,segura20111st,segura2013semeval,krallinger2017overview}. In these tasks, relevant biomedical entities such as genes, proteins and chemicals are given and the information extraction methods aim to identify relations alone or relations and events together within a sentence span. 

A variety of methods have been evaluated on these tasks, which range from rule based methods to more complex machine learning methods, either supported by shallow or deep learning approaches.
Some of the deep learning based methods include CNNs \cite{bjorne2018biomedical,santos2015classifying,zeng2015distant} and RNNs \cite{li2019biomedical,mallory2015large,verga2015multilingual,zhou2016attention}. CNNs will identify local context relations while their performance may suffer when entities need to be identified in a broader context. On the other hand, RNNs are difficult to parallelize while they do not fully solve the long dependency problem \cite{verga2018simultaneously}. Moreover, such approaches are proposed for relation extraction, but not to extract nested events. In this work, we intend to improve over existing methods. We combine a set of parallel multi-head attentions with a set of 1D convolutions to provide global locality in biomedical event and relation extraction. Our approach models locality for self-attention while the interactions between features are learned by multi-head attentions. We evaluate our model on data from the shared tasks for BioNLP 2009, 2011 and 2013, and BioCreative 2017.

The BioNLP Event Extraction tasks provide the most complex corpora with often large sets of event types and at times relatively small corpus sizes. Our proposed approach achieves higher performance on the GE09, GE11, EPI11, ID11, REL11, GE13, CG13 and PC13 BioNLP Shared Task corpora, compared to the top performing system (TEES) \cite{bjorne2018biomedical} for both relation and event extraction in these tasks.  Since the annotations for the test sets of the BioNLP Shared Task corpora are not provided, we uploaded our predictions to the task organizers’ servers for evaluation. 

The CHEMPROT corpus in the BioCreative VI Chemical–Protein relation extraction task (CP17) also provides a standard comparison with current methods in relation extraction. The CHEMPROT corpus is relatively large compared to its low number of five relation types. Our
model outperforms the best-performing system (TEES) \cite{bjorne2018biomedical} for relation extraction in this task. 

\section{Data}\label{sec:data}

We develop and evaluate our approach on a number of event and relation extraction corpora. These corpora originate from three
BioNLP Shared Tasks \cite{kim2009overview,bjorne2011generalizing,nedellec2013overview} and the BioCreative VI Chemical–Protein relation extraction task \cite{krallinger2017overview}. The BioNLP corpora cover various domains of molecular biology and provide the most complex event annotations. The BioCreative corpora use pairwise relation annotations. Table \ref{tbl:data} shows information about these corpora. 

\begin{table}
	\centering
	\small
	\begin{tabular}{l*{14}{c}}
		\toprule
		\midrule
		\textbf{Corpus} &\textbf{Domain}&\textbf{E}&  \textbf{I}& \textbf{S}\\
		\midrule
		\textsc{GE09} & Molecular Biology& 10 & 6 & 11380 \\
		\midrule
		\textsc{GE11} & Molecular Biology& 10 & 6 & 14958 \\
		\textsc{EPI11} & Epigenetics and PTM:s & 16 & 6 & 11772 \\
		\textsc{ID11} & Infection Diseases & 11 & 7 & 5118 \\
		\textsc{REL11} & Entity Relations & 1 & 2 & 11351 \\
		\midrule
		\textsc{GE13} & Molecular Biology & 15 & 6 & 8369 \\
		\textsc{CG13} & Cancer Genetics & 42 & 9 & 5938 \\
		\textsc{PC13} & Pathway Curation & 24 & 9 & 5040 \\
		\midrule
		\textsc{CP17} & Chemical-Protein Int. & - & 5 & 24594 \\
		\midrule
		\bottomrule
	\end{tabular}
	\caption{Information about the domain, number of event and entity types (E), number of event argument and relation types (I), and number of sentences (S), related to the corpora of the biomedical shared tasks}
	\label{tbl:data}\vspace{-0.2cm}
\end{table}

For further analysis and experiments, we also used the AMIA gene-mutation corpus available in \cite{jimeno2018hybrid}. 
The training/testing sets contain 2656/385 mentions of mutations, and 2799/280 of genes or proteins, and 1617/130 relations between genes and mutations. We extracted about 30\% of the training set as the validation set. 


\section{Model}\label{sec:model}

We propose a new biomedical event extraction model that is mainly built upon multi-head attentions to learn the global interactions between each pair of tokens, and convolutions to provide locality. The proposed neural network architecture consists of 4 parallel multi-head attentions followed by a set of 1D convolutions with window sizes 1, 3, 5 and 7. Our model attends to the most important tokens in the input features\footnote{We choose different embeddings for each task/dataset to be in line with TEES.}, and enhances the feature extraction of dependent elements across multiple heads, irrespective of their distance. Moreover, we model locality for multi-head attentions by restricting the attended tokens to local regions via convolutions. 

The relation and event extraction task is modelled as a graph representation of events and relations \cite{bjorne2018biomedical}. Entities and event triggers are nodes, and relations and event arguments are the edges that connect them. An event is modelled as a trigger node and its set of outgoing edges. Relation and event extraction are performed through the following classification tasks: (i) Entity and Trigger Detection, which is a named-entity recognition task where entities and event triggers in a sentence span are detected to generate the graph nodes; (ii)  Relation and Event Detection, where relations and event arguments are predicted for all valid pairs of entity and trigger nodes to create the graph edges; (iii) Event Duplication, where each event is classified as an event or a negative which causes unmerging in the graph\footnote{Since events are n-ary relations, event nodes may overlap.}; (iv) Modifier Detection, in which event modality (speculation or negation) is detected.
In relation extraction tasks where entities are given, only the second classification task is partially used.

The same network architecture is used for all four classification tasks, with the number of predicted labels changing between tasks.

\subsection{Inputs}

The input is modelled in the context of a sentence window, centered around the target entity, relation or event. The sentence is modelled as a linear sequence of word tokens.
Following the work in \cite{bjorne2018biomedical}, we use a set of embedding vectors as the input features, where each unique word token is mapped to the relevant vector space embeddings. We use the pre-trained 200-dimensional word2vec vectors \cite{mikolov2013distributed} induced on a combination of the English Wikipedia and the millions of biomedical research articles from PubMed and PubMed Central \cite{moen2013distributional}, along with the 8-dimensional embeddings of relative positions, and distances learned from the input corpus. 
Following the work in \cite{zeng2014relation}, we use Distance features, where the relative distances to tokens of interest are mapped to their own vectors. We also consider Relative Position features to identify the locations and roles (i.e., entities, event triggers, and arguments) of tokens in the classified structure. 
Finally, these embeddings with their learned weights\footnote{The only exception is for the word vectors, where the original weights are used to provide generalization to words outside the task’s training corpus.} are concatenated together to shape an n-dimensional vector $e_i$ for each word token. This merged input sequence is then processed by a set of parallel multi-head attentions followed by convolutional layers.

\subsection{Multi-head Attention}

Self-attention networks produce representations by applying attention to each pair of tokens from the input sequence, regardless of their distance. According to the previous work \cite{vaswani2017attention}, multi-head attention applies self-attention multiple times over the same inputs using separately normalized parameters (attention heads) and combines the results, as an alternative to applying one pass of attention with more parameters. The intuition behind this modeling decision is that dividing the attention into multiple heads makes it easier for the model to learn to attend to different types of relevant information with each head. The self-attention updates input embeddings $e_i$ by performing a weighted sum over all tokens in the sequence, weighted by their importance for modeling token $i$. Given an input sequence $E=\{e_1, ..., e_I\} \in \mathbb{R}^{I \times d}$, the model first projects each input to a key $k$, value $v$, and query $q$, using separate affine transformations with ReLU activations \cite{glorot2011deep}. Here, $k$, $v$, and $q$ are each in $\mathbb{R}^\frac{d}{H}$, where $d$ indicates the hidden size, and $H$ is the number of heads. The attention weights $a_{ij}^h$ for head $h$ between tokens $i$ and $j$ are computed using scaled dot-product attention:

\begin{equation}
{a_{ij}^h} = \sigma (\frac{{q_i^h}^Tk_j^h}{\sqrt{d}}) 
\end{equation}\vspace{-0.2cm}
\begin{equation}
o_i^h = \sum_j{v_j^h}\odot s_{ij}^h
\nonumber
\label{eq:matt}
\end{equation}

\noindent where $o_i^h$ is the output of the attention head $h$. $\odot$ denotes element-wise multiplication and $\sigma$ indicates a softmax along the $j$th dimension. The scaled attention is meant to aid optimization by flattening the softmax and better distributing the gradients \cite{vaswani2017attention}. The outputs of the individual attention heads
are concatenated into $o_i$ as: $o_i = [o_i^1; ...; o_i^H]$. Herein, all layers use residual connections between
the output of the multi-headed attention and its input. Layer normalization \cite{lei2016layer}, $LN(.)$, is then applied to the output: $m_i = LN(e_i+o_i)$. The multi-head attention layer uses a softmax activation function.

\subsection{Convolutions}

The multi-head attentions are then followed by a set of parallel 1D convolutions with window sizes 1, 3, 5 and 7. Adding these explicit n-gram modelings helps the model to learn to attend to local features. Our convolutions use the ReLU activation function. We use $C(.)$ to denote a convolutional operator. The convolutional portion of the model is given by:

\begin{equation}
c_i = ReLU(C(m_i))
\label{eq:conv}
\end{equation}

Global max pooling is then applied to each 1D convolution and the resulting features are merged together into an output vector. 

\subsection{Classification}

Finally, the output layer performs the classification, where each label is represented by one neuron. The classification layer uses the sigmoid activation function. Classification is performed as multilabel classification where each example may have zero, one or multiple positive labels. 

We use the \textit{adam optimizer} with \textit{binary crossentropy} and a learning rate of 0.001. Dropout of 0.1 is also applied at two steps of merging input features and global max pooling to provide generalization. 

\section{Experiments and Results}\label{sec:expr}

\begin{table*}[h!]
	\centering
	\small
	\setlength{\tabcolsep}{18pt}
	\begin{tabular}{l*{14}{c}}
		\toprule
		\midrule
		\textbf{Task} &\textbf{Precision}&\textbf{Recall}&\textbf{F-score}&\textbf{Approach}\\
		\midrule
		& \underline{65.73} & 44.72 & 53.23&TEES 4CNN\\ 
		&65.01 & \textbf{46.83} & \textbf{54.44}&Proposed 4MHA\\
		\textbf{\textsc{GE09}}&64.37 & 45.19 & 53.10&Proposed 1MHA\\
		&61.99 & 45.51 & 52.48&Proposed 4CNN-4MHA\\
		&\textbf{65.98} &  \underline{45.60} & \underline{53.93}&Proposed 4MHA-4CNN\\
		\midrule
		&66.09 & 46.62 & 54.68&TEES 4CNN\\ 
		&66.19 &  \underline{48.67}  & \underline{56.09}&Proposed 4MHA\\
		\textbf{\textsc{GE11}}& \underline{66.26} & 48.60 &56.07&Proposed 1MHA\\
		&\textbf{67.07} & 47.61 & 55.69&Proposed 4CNN-4MHA\\
		&66.12 & \textbf{49.34} & \textbf{56.51}&Proposed 4MHA-4CNN\\
		\midrule
		&63.31 & 46.73 & 53.78&TEES 4CNN\\ 
		&63.71 & \textbf{50.73}   & \underline{56.48}&Proposed 4MHA\\
		\textbf{\textsc{EPI11}}&\textbf{66.38}&  \underline{49.85} & \textbf{56.94}&Proposed 1MHA\\
		&63.60 & 45.72   & 53.20&Proposed 4CNN-4MHA\\
		& \underline{65.43} & 48.55   & 55.74&Proposed 4MHA-4CNN\\
		\midrule
		& \underline{70.14} & 44.36 & 54.35&TEES 4CNN\\ 
		&66.63 & \textbf{48.65} &  \underline{56.24}&Proposed 4MHA\\
		\textbf{\textsc{ID11}}&\textbf{71.64} &  \underline{46.99}  &\textbf{ 56.75}&Proposed 1MHA\\
		&68.92 & 41.04 & 51.44&Proposed 4CNN-4MHA\\
		&69.05 & 44.91 & 54.43&Proposed 4MHA-4CNN\\
		\midrule
		&71.26 & 62.37 & 66.52&TEES 4CNN\\ 
		& \underline{71.56}& 63.78 &\underline{67.45}&Proposed 4MHA\\
		\textbf{\textsc{REL11}}&68.55 & \underline{64.39} & 66.40&Proposed 1MHA\\
		&71.02 &55.53 & 62.33&Proposed 4CNN-4MHA\\
		&\textbf{71.91}& \textbf{65.39} & \textbf{68.50}&Proposed 4MHA-4CNN\\
		\midrule
		&\textbf{62.22}&39.96&\underline{48.66}&TEES 4CNN\\ 
		& \underline{60.68}&40.35&48.47&Proposed 4MHA\\
		\textbf{\textsc{GE13}}&60.21& \underline{40.75}&48.60&Proposed 1MHA\\
		&58.14&37.66&45.71&Proposed 4CNN-4MHA\\
		&59.76&\textbf{41.65}&\textbf{49.09}&Proposed 4MHA-4CNN\\
		\midrule
		&66.08 & 49.05 & 56.30&TEES 4CNN\\ 
		& \underline{65.92} & \textbf{53.50} &  \textbf{59.06}&Proposed 4MHA\\
		\textbf{\textsc{CG13}}&\textbf{67.02} &  \underline{52.49} &  \underline{58.87}&Proposed 1MHA\\
		&61.91 & 48.02  & 54.09&Proposed 4CNN-4MHA\\
		&65.47 & 51.71  &  57.78&Proposed 4MHA-4CNN\\
		\midrule
		&\textbf{63.49} & 43.37 & 51.54&TEES 4CNN\\
		&59.45 & \textbf{49.90} & \underline{54.26}&Proposed 4MHA\\
		\textbf{\textsc{PC13}}& \underline{60.64} & 47.25  & 53.11&Proposed 1MHA\\
		&57.61 & 43.23 & 49.39&Proposed 4CNN-4MHA\\
		&60.51 &  \underline{49.43} &  \textbf{54.41}&Proposed 4MHA-4CNN\\
		\midrule
		&73.00  & 45.00  & 56.00&TEES 4CNN\\
		&70.00  & \textbf{58.00}   & \textbf{63.00}&Proposed 4MHA\\
		\textbf{\textsc{CP17}}&\textbf{77.00}   &  48.00  &  58.00&Proposed 1MHA\\
		&\textbf{77.00}   & 44.00  & 56.00&Proposed 4CNN-4MHA\\
		&75.00   & \underline{50.00}   & \underline{60.00}&Proposed 4MHA-4CNN\\
		\midrule
		&84.41 &87.52 & 85.90&TEES 4CNN\\
		&83.73&88.51&86.01&Proposed 4MHA\\
		\textbf{\textsc{AMIA}}& \underline{85.12} & \underline{89.50}&\underline{87.31}&Proposed 1MHA\\
		&85.02&89.01&87.00&Proposed 4CNN-4MHA\\
		&\textbf{85.21}&\textbf{90.11}&\textbf{87.53}&Proposed 4MHA-4CNN\\
		\midrule
		\bottomrule
	\end{tabular}
	\caption{Precision, Recall and F-score, measured on the corpora of various shared tasks 
		for our models, and the state of the art. The best scores (the first and the second highest scores) for each task are bolded and highlighted, respectively. All the results (except those of CP17 and AMIA) are evaluated using the official evaluation program/server of each task.}
	\label{tbl:res}
\end{table*}

\begin{figure*}[h!]
	\centering
	\scriptsize
	\setlength{\tabcolsep}{18pt}
	\begin{tabular}{c*{14}{c}}
		\includegraphics[scale=0.32]{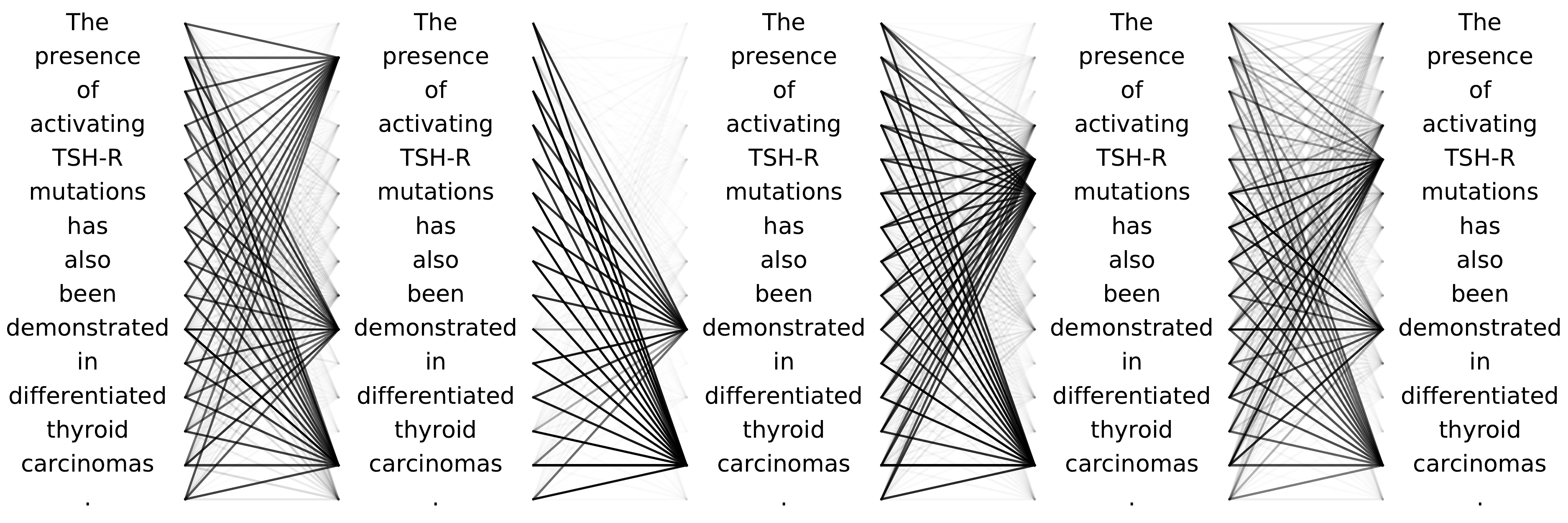}\hspace{-1.0cm} &\vline& \hspace{-2.5cm}\includegraphics[scale=0.32]{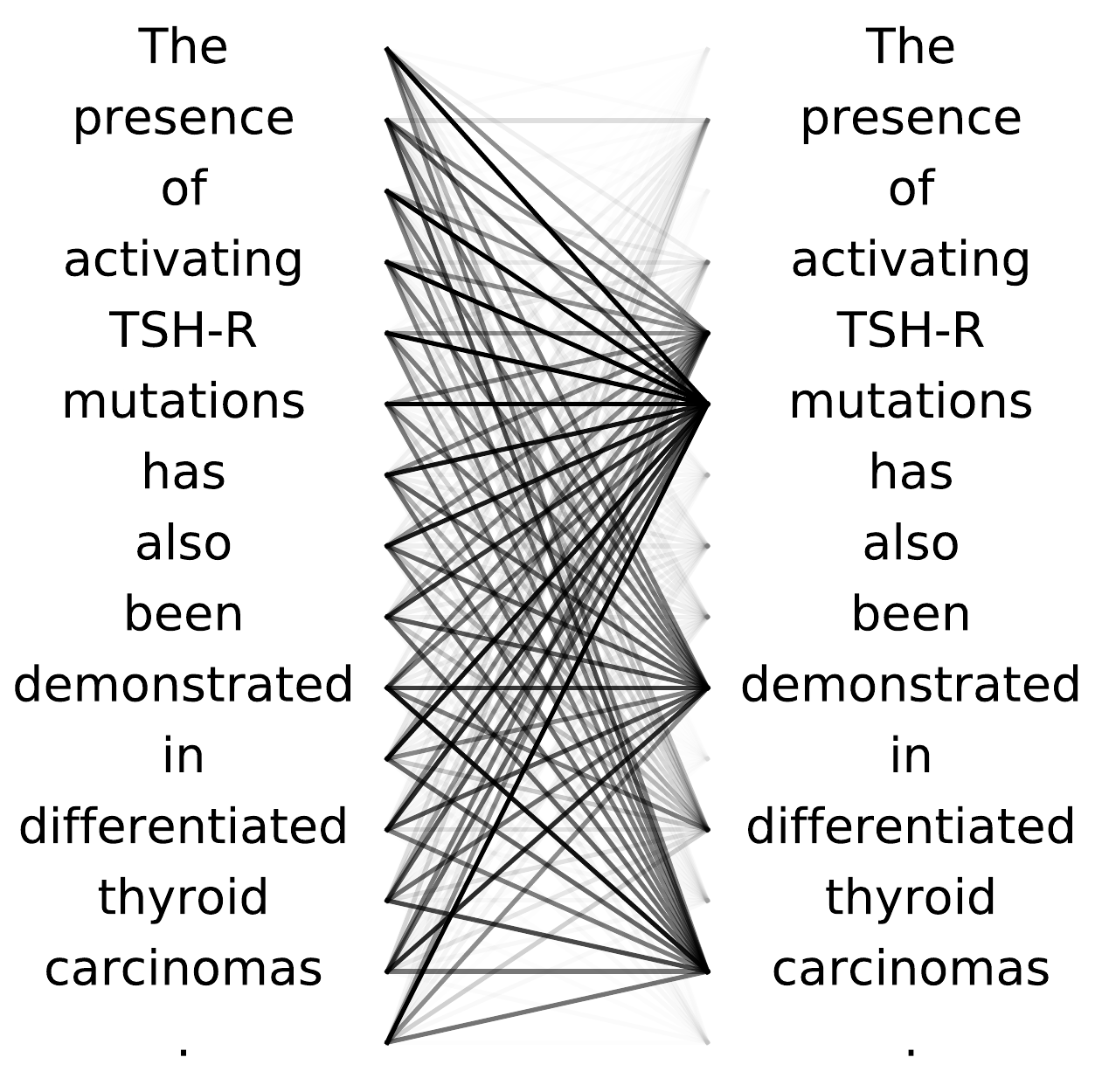}\\
		(4MHA-4CNN)\hspace{-1.0cm}&&\hspace{-1.0cm}(1MHA)\\
		\\
		\includegraphics[scale=0.32]{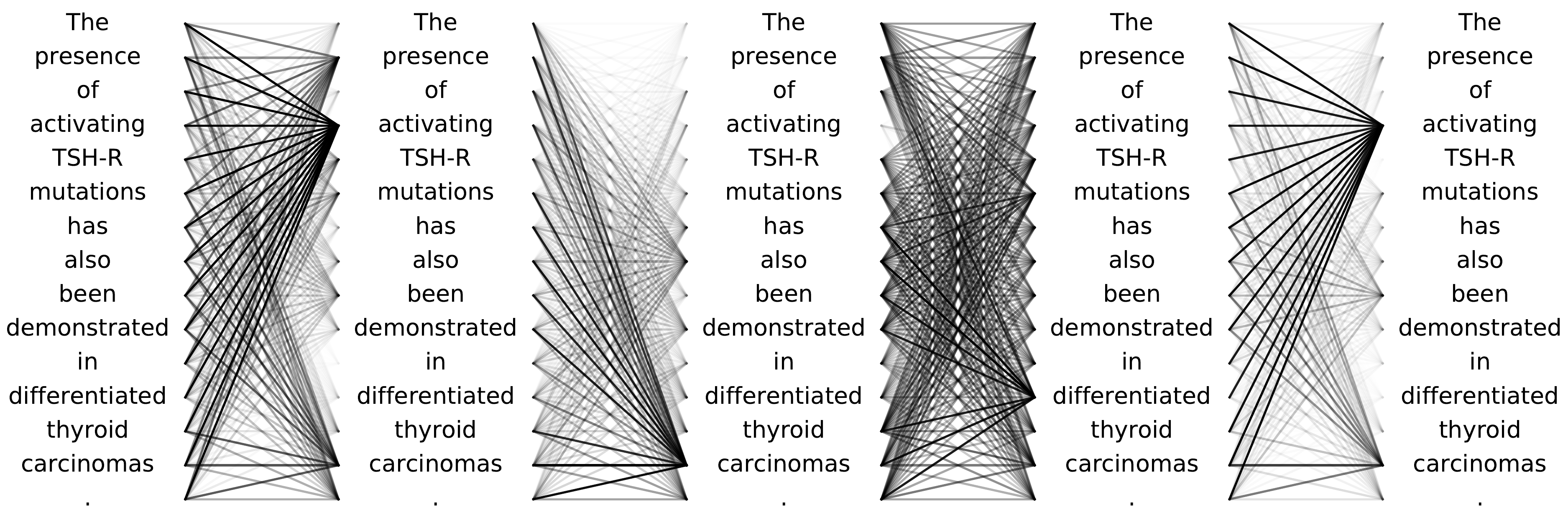}\hspace{-1.0cm} & &\\
		(4MHA)&\\
	\end{tabular}
	\caption{Visualization of multi-head attention in different architectures}
	\label{fig:att}
\end{figure*}

We have conducted a set of experiments to evaluate our proposed approach over the benchmark biomedical corpora. In addition to evaluating our main model (4MHA-4CNN), we have evaluated the performance of three variants of our proposed approach: (i) 4MHA: 4 parallel multi-head attentions apply self-attention multiple times over the input features; (ii) 1MHA: only 1 multi-head attention applies self-attention to the input features; (iii) 4CNN-4MHA: multiple self-attentions are applied to the input features via a set of 1D convolutions\footnote{We also conducted experiments with 1CNN-1MHA and 1MHA-1CNN, which are excluded due to the poor performance.}.  The 4CNN architecture matches the best performing configuration (4CNN - mixed 5 X ensemble)\footnote{We use 4CNN to represent this configuration.} used by TEES \cite{bjorne2018biomedical}, which is composed of four 1D convolutions with window sizes 1, 3, 5 and 7. In our models and TEES, we set the number of filters for the convolutions to 64. The number of heads for multi-head attentions is also set to 8. The reported results of TEES are achieved by running their out-of-the-box system for different tasks. 

Since training a single model can be prone to overfitting if the validation set is too small  \cite{bjorne2018biomedical}, we use mixed 5 model ensemble, which takes 5-best models (out of 20), ranked with micro-averaged F-score on randomized train/validation set split, and considers their averaged predictions. These ensemble predictions are calculated for each label as the average of all the models’ predicted confidence scores. Precision, recall, and F-score of the proposed approach and its variants are compared to TEES in Table \ref{tbl:res}. Our model (4MHA-4CNN) obtains the state-of-the-art results compared to those of the top performing system (TEES) in different shared tasks: BioNLP (GE09, GE11, EPI11, ID11, REL11, GE13, CG13, PC13), BioCreative (CP17), and the AMIA dataset. 

Analyzing the results, we observe that the proposed 4MHA-4CNN model has the best F-score in the majority of datasets except for EPI11, ID11 and CG13, where the proposed MHA models (i.e., 1MHA and 4MHA) have the best F-score and recall. These tasks are related to epigenetics and post-translational modifications (EPI11), infection diseases (ID11) and cancer genetics (CG13), where events typically require long dependencies in most of the cases. It explains why the MHA-alone models are better than when combined with convolutions. The F-scores achieved by 4MHA-4CNN and 4MHA models on GE09 dataset are also very close. In many cases, when using the configurations in which MHA is applied to the input features, both precision and recall are better compared to other configurations. Moreover, having four parallel MHAs applied to the input features outperforms 1MHA and the other potential variants\footnote{The experiment with 8MHA, and multiple MHAs one after the other on the whole sequence are excluded from the paper due to the poor perfromance.}.  

In terms of precision, the advantage of applying 4CNN versus 4MHA to the merged input features depends on the dataset. On PC13, the precision when using 4CNN on the merged input features is much higher compared to other configurations, but the recall is significantly lower. 

The proposed 4MHA-4CNN model has also good recall, except for EPI11, ID11, and CG13, where 4MHA is better. As mentioned before, the addition of convolutions after the multi-head attentions might be less useful in these three sets, since sentences in these topics describe interactions for which long context dependencies are present.

Overall, our observations support the hypothesis that higher recall/F-score is obtained in configurations in which 4MHA is applied first to the merged input features, where CNNs are not as convenient as MHAs to deal with long dependencies.

\begin{figure*}[h!]
	\centering
	\includegraphics[scale=0.45]{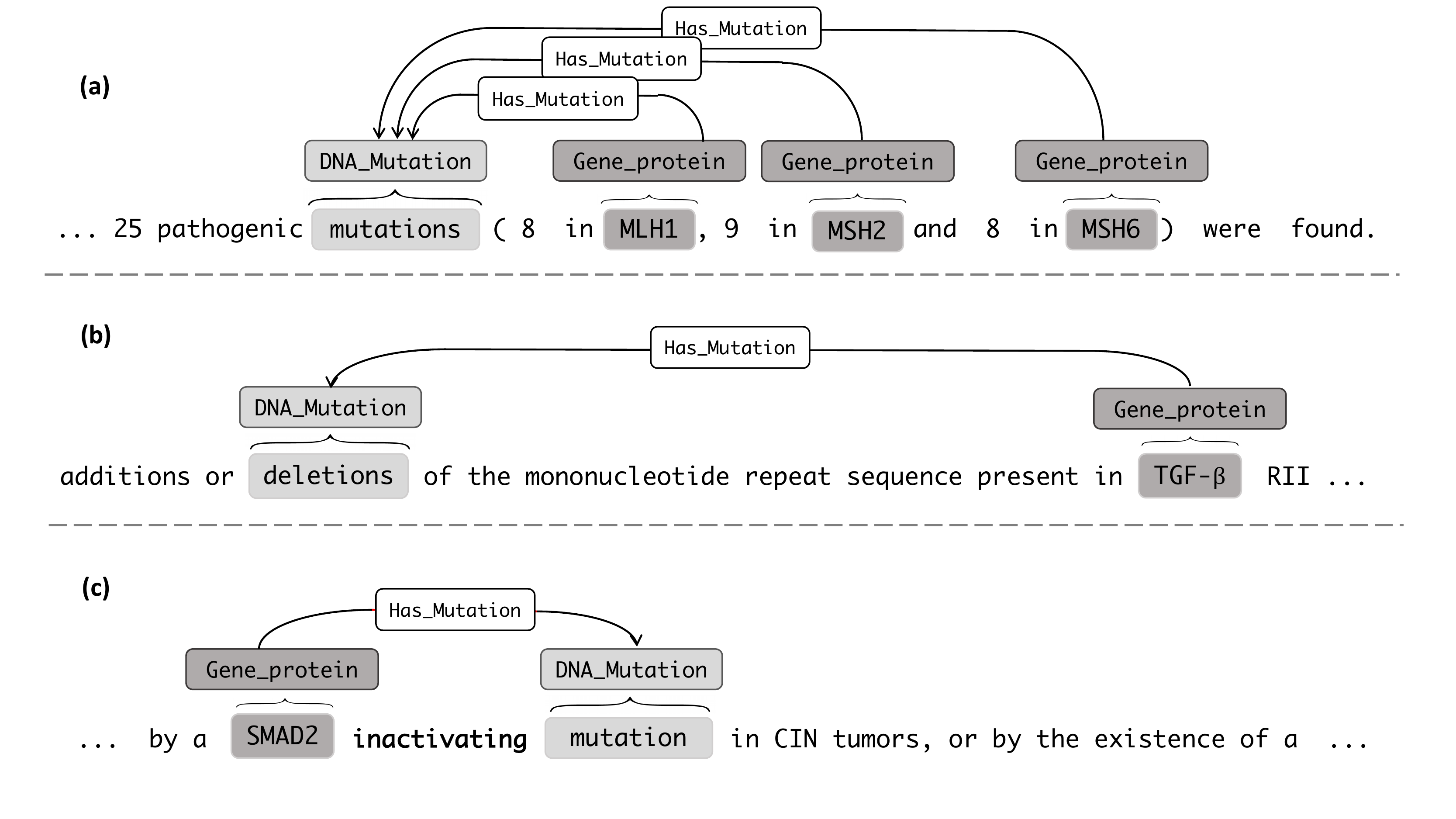}
	\caption{Error analysis of TEES and our approach over the gene-mutation AMIA dataset}
	\label{fig:brt}
\end{figure*}

\subsection{Discussion}

Besides improving the previous state of the art, the results indicate that combining multi-head attention with convolution provides an effective performance compared to individual components. Among the variants of our model, 4MHA also outperforms TEES over all the shared tasks reported in Table \ref{tbl:res}. Even though convolutions are quite effective \cite{bjorne2018biomedical} on their own, multi-head attentions improve their performance being able to deal with longer dependencies.

Figure~\ref{fig:att} shows the multi-head attention (sum of the attention of all heads) of the "relation and event detection" classification task for different proposed network architectures (4MHA-4CNN, 1MHA, and 4MHA) on a sample sentence \textit{"The presence of activating TSH-R mutations has also been demonstrated in differentiated thyroid carcinomas."}. In the 4MHA and 4MHA-4CNN models, the four multi-head attention layers contribute distinctively different attentions from each other. This allows the 4MHA and 4MHA-4CNN models to independently exploit more relationships between the tokens than the 1MHA model. In addition, the convolutions make the 4MHA-4CNN model have more focused attentions on certain important tokens than the 4MHA model. 

Considering the computational complexity, according to the work in \cite{vaswani2017attention}, self-attention has a cost that is quadratic with the length of the sequence, while the convolution cost is quadratic with respect to the representation dimension of the data. The representation dimension of the data is typically higher compared to the length of individual sentences. Outperforming convolutions in terms of computational complexity and F-score, multi-head attention mechanisms seem to be better suited. Although the addition of convolutions after the multi-head makes the model more expensive, the lower representation dimension of the filters reduces the cost. 

\begin{table*}[h]
	\begin{tabular}{ccc}
		\hspace{-0.3cm}\includegraphics[scale=0.42]{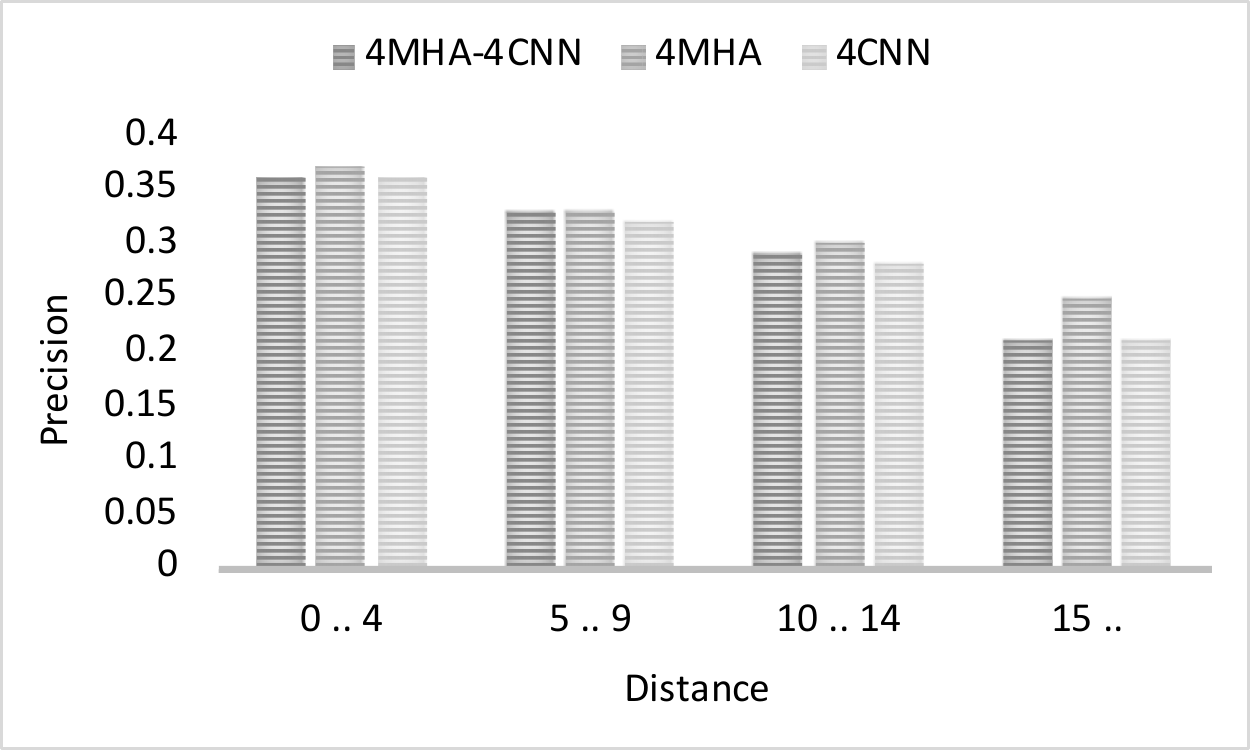}\hspace{-0.35cm} &
		\includegraphics[scale=0.42]{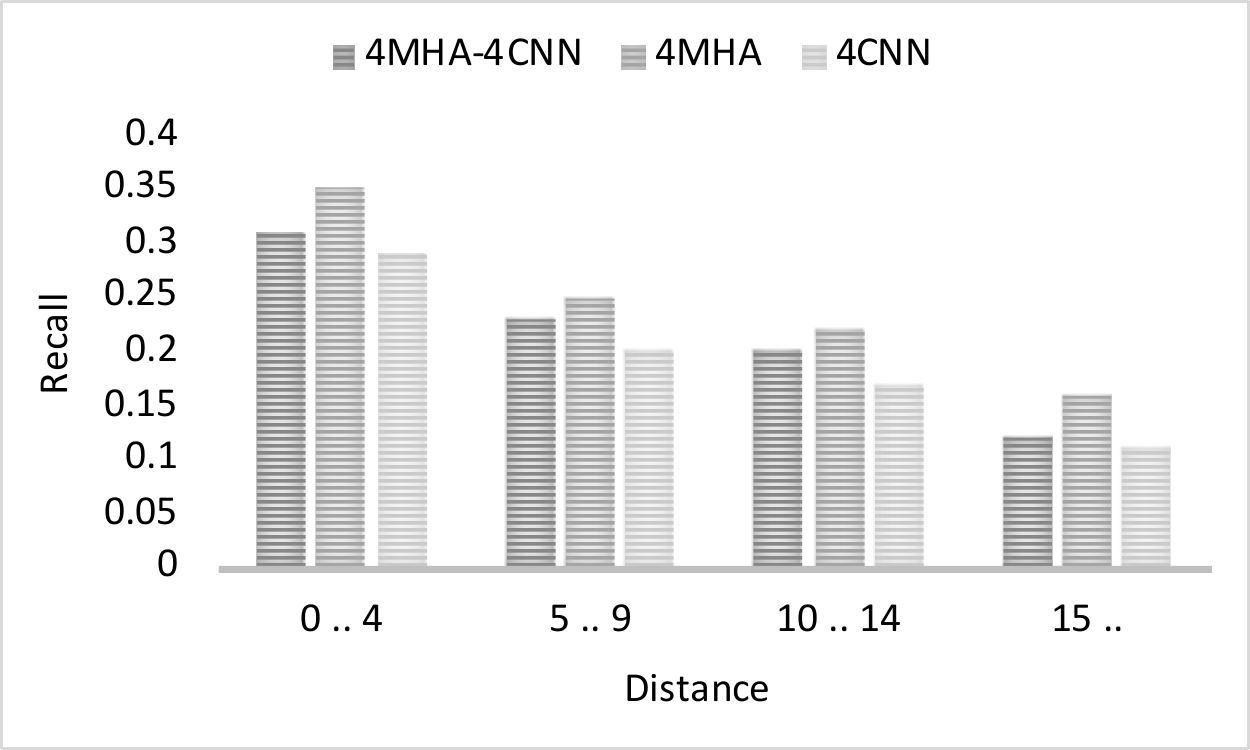}\hspace{-0.35cm} & 
		\includegraphics[scale=0.42]{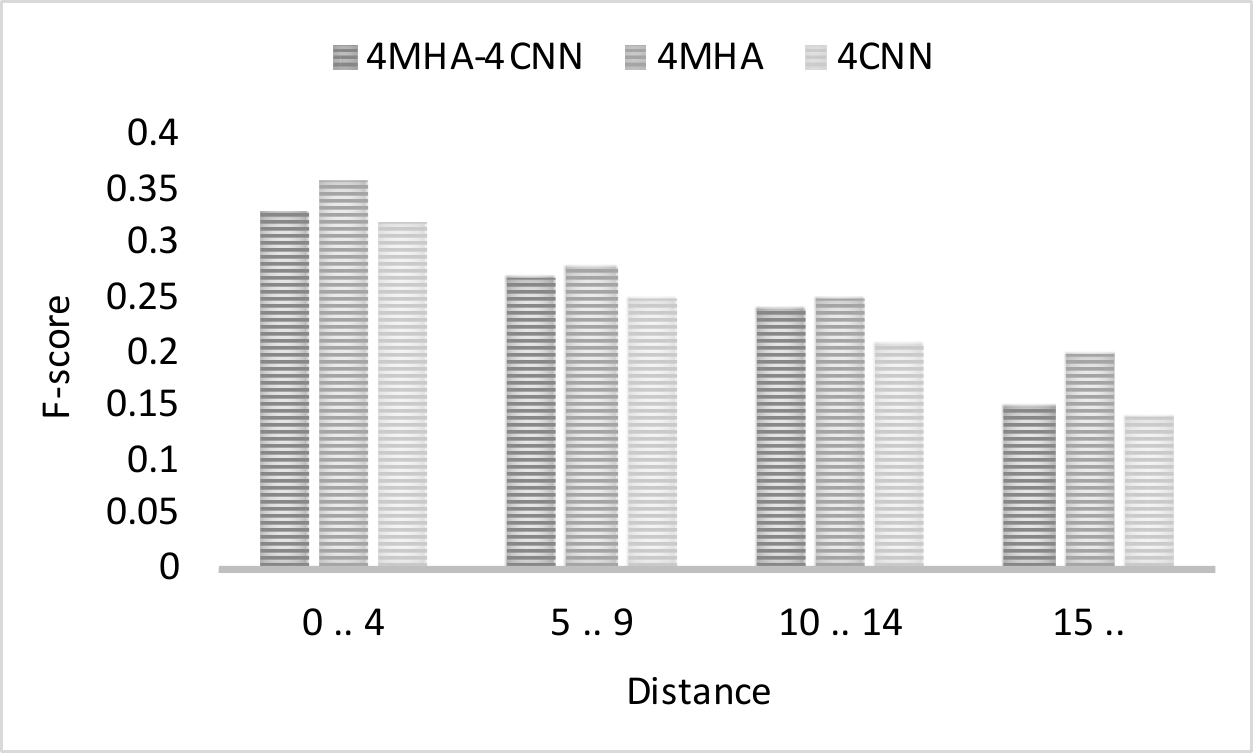}\\
	\end{tabular}
	\caption{Empirical evaluation of long-distance dependencies on CP17}
	\label{fig:dist}
\end{table*}

\subsection{Error Analysis}

We have performed error analysis on the baseline system (TEES), and our approach\footnote{We consider the same configuration for the convolutions in both TEES and our approach.} over the gene-mutation AMIA and CP17 datasets\footnote{We only use these datasets for error analysis due to the limited access to the gold set of other datasets. Hence, this error analysis only covers relation extraction.}, and observed the following sources of error. 

\paragraph{Relations involving multiple entities:} This is a major source of false negatives for TEES, while our approach exhibits a more robust behavior and achieves full recall. The reason would be the ability of multi-head attention to jointly attend to information from different representation subspaces at different positions \cite{vaswani2017attention}.  In an example from the AMIA dataset (Figure \ref{fig:brt} (a)), there is a "has\_mutation" relationship between the term "mutations" and the three gene-protein entities of "MLH1", "MSH2", and "MSH6". While the state-of-the-art approach only finds the relation between the mutation and the first gene-protein (MLH1) and ignores the other two relations, our approach captures the relations between the mutation and all three entities (MLH1, MSH2, and MSH6).

\paragraph{Long-distance dependencies:} TEES also seems to have difficulty in annotating long-distance relations, as in the missed relation between "deletions" and "TGF-$\beta$" in an example from the AMIA dataset (Figure \ref{fig:brt} (b)), which is captured by our approach. We explored this issue further by plotting the performance of different proposed architectures and that of TEES over different distances. We relied on the CP17 dataset, since the test set is considerably larger than AMIA. We performed this analysis for the best performing network architecture proposed (4MHA-4CNN) along with 4MHA and 4CNN architectures separately as the individual components, to study how these architectures behave in capturing distant relations. We measure the distance as the number of tokens between the farthest entities involved in a relation, by employing the tokenization carried out by the TEES pre-processing tool. The results are provided in Figure \ref{fig:dist}. Regardless of the evaluation metric used, we observe that the scores decrease at longer distances, and 4MHA outperforms the other two architectures, which lies in the ability of multi-head attention to capture long distance dependencies. This experiment shows how 4MHA provides globality in 4MHA-4CNN, which slightly outperforms 4CNN in longer distances. 

%
%

\paragraph{Negative or speculative contexts:} Regarding the false positives for TEES that are generally well handled by our system, the annotation of speculative or negative language seems to be problematic. For instance, as depicted in Figure \ref{fig:brt} (c), TEES incorrectly captures the relation between "mutation" and "SMAD2", despite the negative cue, "inactivating". Even though our approach correctly ignores this false positive in the short distance, it still captures speculative long dependencies, which motivates a natural extension of our work in future.

\section{Conclusion}\label{sec:concl}

We have proposed a novel architecture based on multi-head attention and convolutions, which deals with the long dependencies typical of biomedical literature. The results show that this architecture outperforms the state of the art on existing biomedical information extraction corpora. While multi-head attention identifies long dependencies in extracting relations and events, convolutions provide the additional benefit of capturing more local relations, which improves the performance of existing approaches. The finding that CNN-before-MHA is outperformed by MHA-before-CNN is very interesting and could be used as a competitive baseline for future work. 

Our ongoing work includes generalizing our findings to other non-biomedical information extraction tasks. Current work is focused on event and relation extraction from a single short/long sentence; we would like to experiment with additional contents to study the behaviour of these models across sentence boundaries \cite{verga2018simultaneously}. Finally, we intend to extend our approach to deal with speculative contexts by considering more semantic linguistic features, e.g., sense embeddings \cite{rothe2015autoextend} on biomedical literature.

\bibliography{anthology,acl2020}
\bibliographystyle{acl_natbib}

\appendix

\end{document}